\documentclass{llncs}

\usepackage{latexsym}
\def\colc{{\sf c}}
\def\cols{{\sf s}}
\def\colp{{\sf p}}
\def\colavar{{\sf A}}
\def\colbvar{{\sf B}}

\newcommand{\bc}{\begin{center}}
\newcommand {\ec}{ \end{center}}
\newcommand {\be} {\begin{enumerate}}
\newcommand {\ee} {\end{enumerate}}
\newcommand {\bi} {\begin{itemize}}
\newcommand {\ei} {\end{itemize}}
\newcommand {\ba} {\begin{array}}
\newcommand {\et} {\end{tabular}}
\newcommand {\bt} {\begin{tabular}}
\newcommand {\ea} {\end{array}}
\newcommand {\bex} {\begin{example}}
\newcommand {\eex} {\end{example}}
\newcommand {\bsex} {\begin{alphex}}
\newcommand {\esex} {\end{alphex}}

\newcommand {\txt}[1]{\mbox{{ \it{#1}}}}
\def\wff{\txt{wff}}

\newcommand {\ra}{\rightarrow}

\def\Covars{{\cal C\kern-.2ex V}}

\def\evnup{\@ifnextchar[{\@evnup}{\@evnup[0pt]}}
\def\@evnup[#1]#2{\setbox1=\hbox{#2}%
\dimen1=\ht1 \advance\dimen1 by -.5\baselineskip%
\advance\dimen1 by -#1%
\leavevmode\lower\dimen1\box1}

\begin{document}

\title{Sloppy identity }

\date{\today}

\author{Claire Gardent}

\institute{Computerlinguistik\\ Universit\"at des Saarlandes \\ D-66041
  Saarbr\"ucken \\
E-mail: claire@coli.uni-sb.de}

\maketitle

\begin{abstract}
  Although sloppy interpretation is usually accounted for by theories
  of ellipsis, it often arises in non-elliptical contexts. In this
  paper, a theory of sloppy interpretation is provided which captures
  this fact. The underlying idea is that sloppy interpretation results
  from a semantic constraint on parallel structures and the theory is
  shown to predict sloppy readings for deaccented and paycheck
  sentences as well as relational-, event-, and one-anaphora.
  It is further shown to capture the interaction of sloppy/strict
  ambiguity with quantification and binding. Finally, it is compared
  with other approaches to sloppy identity, in particular
  \cite{DalShiPer:eahou91,Hardt:diovpe96} and \cite{FieMay:iai94}.
\end{abstract}

\section{Introduction}
\label{s1}

Sloppy interpretation involves two clauses: a {\bf source} or
antecedent clause and a {\bf target} clause that is, a clause containing
a proform (the {\bf target proform}). When the antecedent of the
target proform also contains a proform (the {\bf source proform}), a
sloppy interpretation may arise. In this case, the interpretation of
the target proform differs from the interpretation of its source
antecedent in the interpretation of the source proform. For instance,
in:

\begin{example}
\label{e18}
Jon\/$^1$ [washes his$_1$ car]\/$^2$. Peter does\/$_2$ too.  
\end{example}

\noindent
the VP-ellipsis {\it does} has, among others, the sloppy
interpretation {\it washes Peter's car} (indeed, this is the preferred
reading). The interpretation is sloppy because whereas in the source
clause {\it Jon washes his car} the source proform {\it his} is
interpreted as {\it Jon}, in the target clause {\it Peter does too},
it is re-interpreted as {\it Peter}.

Although it is most often associated with VP-ellipsis, the phenomenon
of sloppy identity is in fact very pervasive, and can occur in a wide
range of configurations for instance: deaccenting (example 2 where the
deaccented material is in bold face), paycheck sentences (example 3),
VPE as a source proform (examples 4 and 5), non-pronominal referential
elements involving implicit arguments (examples 6 and 7), event
anaphora (examples 8 and 9) and {\it one}-anaphora (example
10)\footnote{The material occuring between brackets represents the
  interpretation being considered, in this case, the sloppy
  interpretation. Most examples are from \cite{Kehler:icfitcodi95}.}.

\begin{example}
Jon$^1$ took his$_1$ wife to the station. No, BILL {\bf took his wife to
the station}. {\it (Bill took Bill's wife to the station)}
\end{example}

\begin{example}
\label{e21}
Jon\/$^1$ spent [his\/$_1$ paycheck]\/$^2$ but Peter saved it\/$_2$.  
 {\it (Peter saved Peter's paycheck)}
\end{example}

\begin{example}
\label{e31}
 I'll [help you]\/$^1$ if you [want me to$_1$]\/$^2$. I'll kiss you
 even if you  don't$_2$.  
 {\it (I'll kiss you even if you don't want me to kiss you)}
\end{example} 
\begin{example} 
When Harry [drinks]\/$^1$, I always conceal [my belief that he
shouldn't\/$_1$]\/$^2$. When he gambles, I can't conceal it$_2$.   
 {\it (When he gambles, I can't conceal my belief that he shouldn't gamble)}
\end{example}

\begin{example}
Jon went to a local bar to watch the Super-bowl, and Bob did too.
{\it (Bob went to a bar local to Bob)}
\end{example} \begin{example}
George drove to the nearest hospital, and Fred did too.
{\it (Fred drove to the hospital nearest to Fred)}
\end{example}

\begin{example}
Jon$^1$ got shot by his$_1$ father. That happened to Bob too.
{\it (Bob got shot by Bob's father)}
\end{example} \begin{example}
Jon$^1$ kissed his$_1$ wife, and Bill followed his example (Dahl, 1972)
{\it (Bill kissed Bill's wife)}
\end{example}

\begin{example}
Although Jon$^1$ bought a picture of his$_1$ son, Bill snapped one himself.
{\it (Bill snapped a picture of Bill's son)}
\end{example}
In short, sloppy interpretation can result from many combinations of
source and target proform.  Nevertheless most theories of sloppiness
are restricted either to VPE or to Paycheck Pronouns thus failing to
capture this obvious generalisation.

What are the constraints on sloppy interpretation? We claim that
parallelism plays a fundamental role in triggering sloppiness.
Specifically, it seems that sloppy interpretation is only possible
when the antecedent of the source proform has a parallel counterpart
in the target clause (cf. \cite{DalShiPer:eahou91}).  The following
examples illustrate this. In example (\ref{e4}), {\it Jon} has a
parallel counterpart in the target clause and consequently, a sloppy
interpretation is possible. This sloppy interpretation is however not
available in (\ref{e5}) where {\it Bill} has no parallel counterpart.
\begin{example}
\label{e4}
The policeman who arrested Bill$^1$ [forgot to read him$_1$
his$_1$ rights]\/$^2$ and so did$_2$ the policeman who arrested Jon.
\end{example}
\begin{example}
\label{e5}
The policeman who arrested Bill$^1$ [forgot to read him$_1$ his$_1$ rights]$^2$
and so did$_2$ Peter.
\end{example}
Of course, sloppy/strict ambiguity is not systematic in that
interaction with other linguistic phenomena may block one or the other
of the two possible readings. For instance in (\ref{e10}), interaction with
quantification results in ruling out a strict reading. Similarly, in
example (\ref{e11}), interaction with binding excludes a
strict reading. Finally, example (\ref{e12}) shows that the
sloppy/strict ambiguity is sensitive to the Pronoun/PN distinction:
the source proform must be a pronoun (and not a proper name) for
sloppiness to be possible.
\begin{example}
\label{e10}
Every$^1$ man believes he$_1$ is a fool. No, PETER believes he is a fool.
(sloppy only) \end{example} 
\begin{example}
\label{e11}
Mary persuaded Jon$^1$ to shave himself$_1$. No, 
Mary persuaded PETER to shave himself. (sloppy only)
\end{example}
\begin{example}
\label{e12}
Jon's mother loves Jon and Bill's mother does too (strict only)
\end{example}
In summary, a theory of sloppiness should be general enough to
encompass the various configurations in which sloppy/strict ambiguity
may arise while correctly accounting for the parallelism constraint
illustrated by examples (\ref{e4}) and (\ref{e5}). Furthermore, it
should predict the interaction of sloppy/strict ambiguity with such
independent phenomena as quantification, binding theory and syntactic
categorisation. 

In this paper, I present a theory of sloppiness which adheres to these
requirements. The proposed analysis is based on a simple parallelism
constraint which is claimed to govern the interpretation of parallel
propositions. In essence, this constraint requires that parallel
structures share a non-trivial part of their semantics.  Crucially,
this constraint is stated in terms of equations and solved using
Higher-Order Unification. It is the use of this particular mechanism
which allows us to make the appropriate linguistic predictions: the
multiple solutions generated by HOU mirror the sloppy/strict of
natural language.

However, HOU is also known to systematically over-generate in that it
yields solutions which although they are mathematically valid, are
linguistically incorrect. To overcome this problem, we use a form of
HOU developed for guiding inductive theorem proving, namely
Higher-Order Coloured Unification (HOCU). The basic idea is that HOCU
provides a general framework in which to model the interface between
semantic construction and other levels of linguistic information. As
we shall see, this yields a linguistically plausible way of avoiding
over-generation.
 
The paper is structured as follows. Section \ref{s2} sketches the
fundamentals of HOCU, and presents the analysis we propose. Section
\ref{s3} shows that this analysis predicts the fact that both ellipsis and
deaccenting may give rise to sloppy/strict ambiguity, section \ref{s4}
that it naturally extends to paycheck pronouns, and section
\ref{s5} that it encompasses those cases of sloppy/strict ambiguity in
which the source proform is a VP-ellipsis. In section \ref{s6}, I review
the remaining sloppy/strict configurations and briefly indicate how
they can be dealt with. Section \ref{s7} focuses on the interaction
of sloppiness with quantification, binding and syntax. Section \ref{s8}
compares the approach with previous proposals and in particular
\cite{Hardt:diovpe96},
\cite{DalShiPer:eahou91} and \cite{FieMay:iai94}.  Section \ref{s9}
concludes with pointers to further research.

\section{The analysis}
\label{s2}

Our analysis of sloppy identity falls out of an independent constraint
on parallel structures. In \cite{Gardent:phouad97}, I argue that this
constraint yields a simple treatment of deaccenting and of its
relationship to ellipsis. In this paper, I show that it also provides
a general theory of sloppy identity.

The parallelism constraint is a semantic constraint which governs the
interpretation of parallel structures. Following
\cite{DalShiPer:eahou91} (henceforth, DSP), we take the structuring
of parallelism as given, that is, we assume that parallel elements are
known. The language used for semantic representation is the typed
lambda calculus and discourse anaphors such as ellipses and
discourse pronouns are represented by free variables of the
appropriate type. For instance, a VP-ellipsis is assigned the semantic
representation $P_{(e,t)}$ that is, a property variable. The
parallelism constraint is as follows.
\begin{quote}
  Given {\it SSem} and {\it TSem}, the semantic representations of the
  source and target utterances and $SP_1 , \dots , SP_n,
  TP_1 , \dots , TP_n$, the semantic representations of the source and
  target parallel elements, the interpretation of parallel utterances
  must obey the following equations: \\

  $ \ba{l} SSem = A(SP_1) ,\dots ,(SP_n) \\ TSem = A(TP_1) ,\dots ,
  (TP_n) \ea $

\end{quote}
Crucially, these equations are resolved using Higher--Order Coloured
Unification. Given the equation $(M = N)$, the unification problem
consists in finding a well-formed coloured substitution of terms for
free variables that will make $M$ and $N$ equal in the theory of
$\alpha\beta\eta$--identity. For instance, given the equation $l(j,m)
= R(j)$, a possible solution is the substitution which assigns
$\lambda x.l(x,m)$ to $R$, written $\{R \leftarrow \lambda x.l(x,m)
\}$. As a result, any free variable occuring in the equations (and in
particular any free variable representing a discourse anaphor) is
assigned a value. As we shall see, this method yields a uniform
account of sloppy/strict ambiguity.

What precisely is Higher-Order Coloured Unification (HOCU) and how
does it differ from standard HOU? For a precise description of HOCU,
the reader is referred to \cite{HuKo:acvotl96}. Briefly, the important
difference is that HOCU operates on a variant of the simply typed
$\lambda$-calculus where symbol occurrences other than bound variables
can be annotated with {\em colours}. Further, colours can be either
constants or variables and restrict the unification process as
follows. Colour variables unify both with colour variables and with
colour constants. In contrast, a colour constant can only unify with
the same colour constant. Further, variables labelled with colour
constants are subject to the following restriction.
\begin{quote}
  For any colour constant \colc\ and any \colc --coloured variable
  $V_\colc$, a well--formed coloured substitution must assign to
  $V_\colc$ a \colc --monochrome term i.e., a term whose symbols are
 $\colc$--coloured.
\end{quote}
Why use HOCU? As we have already shown (cf. \cite{GaKo:hocuanls96}),
the intuition is that colours allows us to create an interface between
semantic construction and other sources of linguistic knowledge. In
this paper, we shall see that it is particularly useful in avoiding
over-generation. 
We now illustrate this point by a simple example. Consider
(\ref{e20}). 
\begin{example}
\label{e20}
Jon$^1$ likes his$_1$ wife. Peter does too.
\end{example}
In an HOU setting, the parallelism constraint yields the following
equations for this discourse (recall that ellipses are represented
using free variables; thus here $R$ represents the ellipsis {\it does
too}):
\begin{center}
$
\begin{array}{rl}
l(j,\txt{wife\_of}(j)) & = A(j) \\
R(p) & = A(p)
\end{array}
$
\end{center}
\noindent
Resolution of the first equation yields a total of 4 solutions,
namely:
\begin{center}
$
\begin{array}{l}
\{A \leftarrow \lambda x.l(x,\txt{wife\_of}(x))\} \\
\{A \leftarrow \lambda x.l(x,\txt{wife\_of}(j))\} \\
\{A \leftarrow \lambda x.l(j,\txt{wife\_of}(x))\} \\
\{A \leftarrow \lambda x.l(j,\txt{wife\_of}(j))\} 
\end{array}
$
\end{center}
Linguistically, only the first two solutions are valid. The last two
are invalid because they would yield an incorrect semantics for the
discourse in (\ref{e20}), namely:
\begin{center}
\begin{tabular}{l}
Jon likes Jon's wife and Jon likes Peter's wife. \\
Jon likes Jon's wife and Jon likes Jon's wife. 
\end{tabular}
\end{center}
Intuitively, the problem is that the term occurrence representing the
source subject appears in the target representation, and that it may
not do so. To capture this intuition, DSP introduce the {\it Primary
  Occurrence Restriction} (POR) which in essence aims at ensuring that
the term(s) representing the source parallel element(s) do not occur in
the solution.
\begin{quote}
{\bf Primary Occurrence Restriction}

  Given a labeling of occurrences as either primary or secondary, the
  POR excludes from the set of linguistically valid solutions, any
  solution which contains a primary occurrence.
\end{quote}
Within the HOCU framework, the POR finds a natural encoding: primary
occurrences are \colp -coloured whilst the free variable representing
an ellipsis is \cols -coloured. Since a well-formed coloured
substitution only assigns to a \colc -coloured free variable (where
\colc\ is a colour constant), a \colc -monochrome term, it follows that
the term assigned to an ellipsis variable ($R_\cols$) may not contain
any primary occurrence (because primary occurrences are \colp
-coloured and \colp $\neq$ \cols). Coming back to example (\ref{e20}),
the HOCU equations are (here and in what follows, we ignore irrelevant
colours; $\colavar, \colbvar$ are colour variables): 
\begin{center}
$
\begin{array}{rl}
l(j_\colp,\txt{wife\_of}(j_\colavar)) & = A_\colbvar(j_\colp) \\
R_\cols(p_\colp) & = A_\colbvar(p_\colp)
\end{array}
$
\end{center}
For which, the following substitutions are well-formed coloured
substitutions
\begin{center}
$
\begin{array}{l}
\{R_\cols \leftarrow \lambda x.l(x,\txt{wife\_of}(x)),
A_\colbvar \leftarrow \lambda x.l(x,\txt{wife\_of}(x)) \}\\
\{R_\cols \leftarrow \lambda x.l(x,\txt{wife\_of}(j_\cols)),
A_\colbvar \leftarrow \lambda x.l(x,\txt{wife\_of}(j_\colavar)) \}\\
\end{array}
$
\end{center}
but not these:
\begin{center}
$
\begin{array}{l}
\{R_\cols \leftarrow  \lambda x.l(j_\colp,\txt{wife\_of}(x)),
A_\colbvar \leftarrow \lambda x.l(j_\colp,\txt{wife\_of}(x)) \}\\
\{R_\cols \leftarrow  \lambda x.l(j_\colp,\txt{wife\_of}(j_\cols)),
A_\colbvar \leftarrow \lambda x.l(j_\colp,\txt{wife\_of}(j_\colavar)) \}\\
\end{array}
$
\end{center}
To summarise: because HOU can yield several solutions (rather than a
single one), it allows us to capture the sloppy/strict ambiguity
displayed in natural language discourse. And because we use HOCU
(rather than straight HOU), we can eliminate from the set of
solutions, those solutions that are linguistically invalid. For more
details on the applications of HOCU to natural language semantics, we
refer the reader to (\cite{GaKo:hocuanls96}). In what follows, we will
omit colours unless they are used for something else than the POR.
Concretely, this means that colours will only re-appear in section
\ref{s7}, where we concentrate on the interaction between semantic
construction and other sources of linguistic information.

\section{Sloppy identity in ellipsis and deaccenting contexts}
\label{s3}

It has often been observed (cf.
\cite{Tancredi:ddap92,ChoLas:papt91,Rooth:erarr92}) that VP-ellipsis
and deaccenting (i.e. prosodic reduction) share a number of
interpretive similarities and in particular that they both give rise
to sloppy/strict ambiguity. Thus in (\ref{e52}), the target VP {\it took his
  wife to the station} is deaccented, and like the elided VP in
(\ref{e51}), it can be interpreted either strictly or sloppily (upper
letters indicate prosodic prominence and bold face deaccented material).
\begin{example}
\label{e51}
Jon$^1$ took his$_1$ wife to the station. No, BILL did. \end{example}
\begin{example} 
\label{e52}
Jon$^1$ took his$_1$ wife to the station. No, BILL {\bf took his wife to
the station.}
\end{example}
Interestingly, this is exactly what the parallelism constraint
predicts. Thus in the VPE case, the parallelism constraint requires
that the following equations hold:
\begin{center}
$
\ba{l}
tk(j, \txt{wife\_of}(j), s) = A(j) \\
R(b) = A(b)
\ea
$
\end{center}
Resolving the first equation yields two possible values for {\it A}:
\begin{center}
$
\ba{l}
\{A \leftarrow \lambda x. tk(x, \txt{wife\_of}(x), s)\} \\
\{A \leftarrow \lambda x. tk(x,
\txt{wife\_of}(j), s) \}
\ea
$
\end{center}
And consequently, the target clause receives two
interpreta\-tions, 
either \\ $tk(b, \txt{wife\_of}(b), s)$ or
$tk(b,\txt{wife\_of}(j),s)$ -- as required. Now consider the
deaccenting case. As for the ellipsis case, we assume that anaphors in
the source are resolved whereas discourse anaphors in the target are
represented using free variables (alternatively, we could resolve them
first and let HOU filter unsuitable resolutions out). Specifically, the target pronoun {\it his} is represented by
the free variable $x$ and the equations to be solved are:
\begin{center}
$
\ba{l}
tk(j, \txt{wife\_of}(j), s) = A(j)\\
tk(b, \txt{wife\_of}(x), s) = A(b)
\ea
$
\end{center}
The first equation is resolved as before, thereby yielding two
possibilities for the second equation:
\begin{center}
$
\ba{l}
tk(b, \txt{wife\_of}(x), s) = tk(b, \txt{wife\_of}(b), s) 
\\

tk(b, \txt{wife\_of}(x), s) = tk(b, \txt{wife\_of}(j), s) 
\ea
$
\end{center}
It follows that $x$ can unify either to $j$ or to $b$ and accordingly,
the target pronoun {\it his} is interpreted as either strict ({\it
  Jon}) or sloppy ({\it Bill}).

Another interesting fact about sloppy interpretation in deaccenting and
VP ellipsis contexts is that it may involve an extended domain of
licensing that is, a domain that extends beyond the clausal level. For
instance, \cite{Rooth:erarr92} notes that in (\ref{e56a}) and
(\ref{e56b}), both the deaccented {\it I was bad-mouthing her} and the
elliptical {\it I was} can be assigned the sloppy interpretation: {\it
I was bad-mouthing Sue} even though {\it Sue} is not part of the
anaphoric clause.
\begin{example}
\label{e56a}
First, Jon told Mary$^1$ I was bad-mouthing her$_1$, and then he told SUE I
was. 
\end{example} 
\begin{example}
\label{e56b}
First, Jon told Mary$^1$ I was bad-mouthing her$_1$, and then he told SUE I
was bad-mouthing her.
\end{example}
More generally, the interpretation of both ellipsis and deaccenting
can depend on elements which occur outwith their governing
categories. This suggests that the semantic licensing of these
two phenomena occurs at the sentential rather than at the clausal level.
Again this falls out of the parallelism constraint: since this
constraint operates on utterance representations, it is not restricted to
clauses but may indifferently apply either to clauses or to
sentences. Specifically, the analysis for the above examples goes as
follows. 

For the ellipsis case, the equations are:
\begin{center}
$
\ba{l}
t(j,m,bm(i,m)) = A(m) 
\\
t(x,s,R(i)) = A(s) 
\ea
$
\end{center}
\noindent
Solving the first equation yields $\lambda z. t(j,z,bm(i,z))$ and
$\lambda z. t(j,z,bm(i,m))$ as possible values for $A$. By
substitution and $\beta$-reduction $A(s)$ is then either
$t(j,s,bm(i,s))$ or $t(j,s,bm(i,m))$ and the resulting equations are
resolved by the following substitutions:
\begin{center}
$
\ba{l}
\{x \leftarrow j, R \leftarrow \lambda z.bm(z,m) \} \\
\{x \leftarrow j, R \leftarrow \lambda z.bm(z,s) \}
\ea
$
\end{center}
where the first solution yields the strict reading, the second the sloppy. 
Resolution of the deaccenting case can similarly be summarised as
follows. This time, the equations resulting from the parallelism
constraint are:
\begin{center}
$
\ba{l}
t(j,m,bm(i,m)) = A(m) 
\\
t(x,s,bm(i,y)) = A(s) 
\ea
$
\end{center}
\noindent
As before, resolving the first equation yields two possible values for
$A$ namely, $\lambda z. t(j,z,bm(i,z))$ and $\lambda z.
t(j,z,bm(i,m))$. Similarly, $A(s)$ is then either \\ $t(j,s,bm(i,s))$
or $t(j,s,bm(i,m))$. Consequently, (\ref{e56b}) also receives both a
strict and a sloppy interpretation.

Finally, deaccenting may involve more complex cases of sloppy
interpretation as illustrated by the following examples. 
\begin{example}
\label{e39b}
Jon$^1$ said he$_1$ was clever. No, PETER said he was intelligent.
\end{example}
\begin{example}
\label{e39c}
The policeman who arrested Jon$^1$ forgot to read him$_1$ his$_1$ rights and so
did the one PETER got collared by.
\end{example}
Example (\ref{e39b}) is a straightforward deaccenting example where
  the deaccented material differs in its lexical realisation from its
  source counterpart; example (\ref{e39c}) is more
  complex and involves both deaccenting and VP-ellipsis. Both cases
  trigger a sloppy/strict ambiguity in the interpretation of the
  target clause.

The important point is that in such cases, ellipsis and deaccenting
differ in the semantic relation they require to hold between source
and target utterances.  In the ellipsis case, the relation is one of
syntactic identity: the semantic representation of the ellipsis in the
target clause must be identical with the semantic representation of
part of the source clause. In the case of deaccenting, the relation is
more subtle.  Consider (\ref{ex:e39a}) for instance.
\begin{example}
\label{ex:e39a}
First, JON called Mary a Republican and then PETER insulted her.
\end{example}
\cite{Rooth:erarr92} argues that in such cases, entailment is involved in
licensing prosodic reduction: (\ref{ex:e39a}) is licensed by the
implication that `if $x$ calls $y$ a republican, then $x$ insults $y$'
and consequently, the deaccented clause is interpreted as {\it Peter
  insulted Mary}. Note that through the entailment, {\it her} is
resolved to {\it Mary}.

However, Rooth himself notes that the licensing relation between
a deaccented sentence and its source, need not always be
entailment. Thus in examples (\ref{e70a}),  (\ref{e70b}) and
(\ref{e70c}), there is clearly no entailment relation holding between
source and target clause.
\begin{example}
\label{e70a}
He bit her and then SHE punched HIM. 
\end{example}
\begin{example}
\label{e70b}
Tell me who assaulted whom? HE bit HER.
\end{example}
\begin{example}
\label{e70c}
First, a policeman arrested Peter and then PAUL got collared by one.
\end{example}
Rather, the requirement seems to be that the two clauses entail a
(reasonably specific) common proposition of a more general form. For
instance, the deaccenting in (\ref{e70c}) seems to be licensed by the
fact that both {\it x arrest y} and {\it x collared y} entail that
{\it x did something nasty to y} (note that if I say {\it First, a
  policeman arrested Peter and then PAUL was invited to TEA by one}, I
have to stress {\it was invited to TEA}). More specifically, one could
follow \cite{Tancredi:ddap92} and argue that deaccenting is licensed
provided $(X_S \supset Y) \wedge (X_T \supset Y)$ where $X_S,X_T$ are
the semantic representations of the source and target clauses
respectively, and $Y$ is such that it subsumes $X_S$ and $X_T$ and it
is entailed by $X_S$ and $X_T$.

Now, whatever the relation is, which holds between a
deaccented clause and its source, the above data is clearly
problematic for the HOU approach. In such cases, the
semantic representation of the deaccented material will differ from
that of its parallel counterpart in the source and therefore HOU
(which is essentially a matching operation on syntactic structures)
will fail thus failing to account for some perfectly acceptable
discourses.

There is an intuitively natural solution to this
problem. As we already argued in \cite{GaKoLe:cahou96,Gardent:phouad97},
a variant of HOU is required to deal with the semantics of deaccenting
namely, HOU augmented with logical relations (HOU+R, cf.
\cite{Kohlhase:hot95}). Crucially, this form of unification takes into
account not only syntactic $\alpha\beta\eta$-identity, but also any
logical relation we care to specify. For instance, if the specified
relation is entailment, the equation $(M = N)$ will be solved by HOU+R
if either $M$ and $N$ syntactically unify or if it can be proved that
$M$ entails $N$.  The point is that whatever the relation is which we
find, holds between a deaccented clause and its source, we can
integrate it into the HOU framework (provided it's a logical
relation). Briefly, the HOU+R approach consists in combining HOU with
theorem proving. Specifically, each time an intermediate equation ($M
= N$) of type $t$ is found, the theorem prover is called to try and
prove that $R(M,N)$ (where $R$ is the specific relation HOU is
incremented with). Thus, if we assume that source and target
clause must share a common property, any attempt to solve $(M = N)$
will trigger the attempt to prove that $(M \supset Y) \wedge (N
\supset Y)$ where $Y$ subsumes both $M$ and $N$. For a more detailed
presentation of HOU+R, we refer the reader to \cite{Kohlhase:hot95};
for a discussion of how this form of unification can be used for
deaccenting, see \cite{GaKoLe:cahou96,Gardent:phouad97}. We now briefly
sketch how such cases can be handled within the HOU+R framework.
Consider example (\ref{e39c}) with sloppy interpretation {\it The
policeman who Peter got collared by forgot to read Peter Peter's
rights}. The equations to be resolved are:
\begin{center}
$
\ba{rl}
\exists x [po(x) \wedge ar(x,j) \wedge f(x,rd(x,j,j'sr))] 
& = A(j) \\
\exists x [Q(x) \wedge col(x,p) \wedge R(x)] 
& = A(p) 
\ea
$
\end{center}
Solving the first equation yields (we concentrate on the sloppy
reading):
\begin{center}
  $ \ba{rl} \{A \leftarrow \lambda z.\exists x [po(x) \wedge ar(x,z)
  \wedge f(x,rd(x,z,z'sr))] \}
\ea $
\end{center}
Consequently, the second equation becomes
\begin{center}
$
\ba{l}
\exists x [Q(x) \wedge col(x,p) \wedge R(x)] 
=
\exists x [po(x) \wedge ar(x,p) \wedge f(x,rd(x,p,p'sr))] 
\ea
$
\end{center}
For which, a possible solution is: $\{Q \leftarrow \lambda z. po(z), R
\leftarrow \lambda z. f(z,rd(z,p,p'sr)) \} $ After substitution, the
lhs of the equation is then: 
\bc
$\exists x [po(x) \wedge col(x,p) \wedge
f(x,rd(x,p,p'sr))]$
\ec
That is, the underspecified semantics of the
target {\it so did the one Peter got collared by} has correctly been
resolved to {\it the policeman who arrested Peter forgot to read Peter
  Peter's rights}.  Note however that the equation is not solved.
Rather, we've reached a situation in which no free variable occurs but
still left and right-hand sides are not identical. Within the HOU+R
setting, an attempt will then be made to prove that R holds between
both sides of the equation. In this case, the theorem prover
must prove that both sides of the equations entail a common more
general proposition. Specifically, it must be proved that:
\begin{center}
$
\ba{l}
\exists x [po(x) \wedge ar(x,p) \wedge f(x,rd(x,p,p'sr)) \supset Y]
\wedge 
\\
\exists x [po(x) \wedge col(x,p) \wedge f(x,rd(x,p,p'sr)) \supset Y]
\ea
$
\end{center}
Since tableaux are refutation systems, we in fact try to prove the
negation of this formula and aim for a contradiction. During the proof
process, each conjunctive formula can be broken down into its
conjuncts which are then added to the current tableau branch.
Conversely, a disjunctive formula triggers a branching of the tree
whereby each new branch is labelled with one of the disjuncts.
Universal and existential formulas license the addition of an
instantiation of this formula to the current branch (whereby the
instantiation drops the quantifier and replaces the bound variable by
an unused free variable in the case of a universal and by a new skolem
term in the case of an existential). Finally, if both $X_1$ and $\neg
X_2$ occur on the same tableau branch, and if a substitution can be
found which makes $X_1$ and $X_2$ equal, the branch is closed.  When
all tableau branches are closed, the theorem is proved (for a more
precise definition of the tableau method, see e.g.
\cite{Fitting:folaat90}).  The tableau for the above example can be
sketched as follows. In a first phase, we get (we abbreviate
$f(x,rd(x,p,p'sr))$ to $fr(x,p)$):

\[\begin{array}{c} 
(1)\txt{   } \forall x\forall y[ar(x,y) \supset nasty(x,y)] \\
(2)\txt{   }  \forall x\forall y [col(x,y) \supset nasty(x,y)] \\
(3)\txt{   }  ar(v_1,v_2) \supset nasty(v_1,v_2) \\
(4)\txt{   }  col(v_3,v_4) \supset nasty(v_3,v_4) \\
(5)\txt{   }  \neg(\exists x [po(x) \wedge ar(x,p) \wedge fr(x,p) \supset Y]
\wedge 
\exists x [po(x) \wedge col(x,p) \wedge fr(x,p) \supset Y]
)
\\
\begin{array}{c|c}
(6)\txt{   }  \neg\exists x [po(x) \wedge ar(x,p) \wedge fr(x,p) \supset Y] &
\neg\exists x [po(x) \wedge col(x,p) \wedge fr(x,p)\supset Y] \\
 \end{array}
 \end{array}\]

 1 and 2 formalise the fact that we assume a context where {\it if x
   arrests y, then x is nasty to y} and further {\it if x collars y,
   then x is nasty to y}. 3 and 4 are from 1 and 2 by instantiation
 (where $v_i$ are new free variables). 5 is the negation of what has
 to be proved. Since $\neg(X \wedge Y)$ is a disjunctive formula, the
 tableau branches and its disjuncts $\neg X$ and $\neg Y$ are added to
 the new branches. We now show the continuing derivation for the left
 branch (the right branch develops in a similar way). For readability,
 we add lines 3 and 6 of the beginning tableau to the continuing tableau.

\[\begin{array}{c} 
(3)\txt{   }ar(v_1,v_2) \supset nasty(v_1,v_2) \\
(6)\txt{   }\neg\exists x [po(x) \wedge ar(x,p) \wedge fr(x,p) \supset Y]  \\
(7)\txt{   }\neg [po(v_1) \wedge ar(v_1,p) \wedge fr(v_1,p)\supset Y] \\
(8)\txt{   }po(v_1 ) \wedge ar(v_1 ,p) \wedge fr(v_1 ,p) \\
(9)\txt{   }\neg Y \\
(10)\txt{   }po(v_1) \\
(11)\txt{   }ar(v_1,p) \\
(12)\txt{   }fr(v_1,p) \\
\begin{array}{c|c}
(13)\txt{   }\neg ar(v_1, v_2) & nasty(v_1,v_2) \\
\{v_2 \leftarrow p \} & \{Y \leftarrow nasty(v_1, v_2) \} \\
\star & \star
\end{array}
\end{array}
\]

7 is from 6 by instantiation. Since $\neg(X\supset Y)$ is a
conjunctive formula with conjuncts $X$ and $\neg Y$, lines 8 and 9 are
added to the tableau (from 7). Similarly, lines 10, 11 and 12 are from
line 8 (because line 8 is labelled with a conjunctive formula).  At
this stage, the disjunctive formula in (3) is used and the tree
branches yielding (13). By using the indicated bindings, we derive the
contradictions $\neg ar(v_1, p) \wedge ar(v_1, p)$ for the leftmost
branch, and $\neg \txt{nasty}(v_1, p) \wedge \txt{nasty}(v_1, p)$ for
the rightmost one.  Since both branches are closed, the tableau is
closed and proposition (1) is proved. Thus HOU+R succeeds thereby
yielding the appropriate sloppy interpretation for the target clause
{\it So did the one Peter got collared by}.

\section{Paycheck Pronouns}
\label{s4}

Why does the parallelism constraint predict paycheck pronouns?
Intuitively, the reason is that as in the ellipsis case, the
$\lambda$-term shared by source and target utterances may contain the
representation of a pronoun.  When the antecedent of this pronoun is a
parallel element, HOU predicts that this pronoun can behave sloppily.
Let us see in more detail how this works. Consider example
(\ref{e21a}).
\begin{example}
\label{e21a}
Jon\/$^1$ spent [his\/$_1$ paycheck]\/$^2$ but Peter saved it\/$_2$.  
\end{example}
The pronoun {\it it} occurring in the second clause has a sloppy
interpretation in that it can be interpreted as meaning {\it Peter's
  paycheck}, rather than {\it Jon's paycheck}. In the literature
such pronouns are known as {\it paycheck pronouns} and are treated as
introducing a definite whose restriction is pragmatically given (cf.
e.g. \cite{Cooper:tiop79}). Within our proposal, we can
straightforwardly capture this intuition by assigning paycheck
pronouns the following representation:
\begin{center}
\bt{llr} 
Pro & $\leadsto
\lambda Q. \exists x[P(x)\wedge\forall y[P(y)
\rightarrow y = x]\wedge Q(x)]$ 
 & with $P \in \wff_{(e \rightarrow t)}$
\et
\end{center}
That is, paycheck pronouns are treated as definites whose restriction
($P$) is a variable of type $(e \ra t)$. Furthermore, we assume that
paycheck pronouns like VP-ellipses, occur in parallel structures and
hence are subject to the parallelism constraint. Under these
assumption the following equations must hold (we abbreviate $\lambda
Q. \exists x[P(x)\wedge\forall y[P(y) \rightarrow y =
x]\wedge Q(x)]$ to$\lambda Q. \exists_1 x[P(x) \wedge Q(x)]$):
\begin{center}
$
\ba{l}
\exists _1x[pc\_of(x,j) \wedge sp(j,x)] = A(j,sp)\\
\exists _1x[P(x) \wedge sa(p,x)] = A(p,sa)
\ea
$
\end{center}
Resolving the first equation yields $\lambda y. \lambda O. \exists
_1x[pc\_of(x,y) \wedge O(y,x)]$ as a value for $A$ so that $A(p,sa) =
\exists _1x[pc\_of(x,p) \wedge sa(p,x)]$ and $\{P \leftarrow \lambda y.
pc\_of(y,p)\}$. That is, the target clause is correctly assigned the
sloppy interpretation: {\it Peter saved Peter's paycheck}.

\section{Source proform: VPE}
\label{s5}
 
So far, we have considered cases of sloppy interpretation in which the
source proform is a pronoun. Interestingly however, the source proform
can also be a VP ellipsis. This is illustrated by the following
examples (from \cite{Hardt:diovpe96}).
\begin{example}
\label{e31a}
 I'll [help you]\/$^1$ if you [want me to$_1$]\/$^2$. I'll kiss you
 even if you  don't$_2$.  
\end{example} 
\begin{example} 
\label{e31b}
When Harry [drinks]\/$^1$, I always conceal [my belief that he
shouldn't\/$_1$]\/$^2$. When he gambles, I can't conceal it$_2$.   
\end{example}
\begin{table}
\label{t1}
\begin{small}
\begin{center}
\begin{tabular}{|l|l|l|}
\hline
& Source Utterance & Target Utterance \\
\hline
&&\\
Configuration &
NP$^j ,\dots ,$ [$_{VP} ,\dots ,$ \fbox{PRO}$_j ,\dots ,$ ]$^i$
& $,\dots ,$ \fbox{VPE}$_i ,\dots ,$ \\
&& \\
Example & {\it Jon$_j$ [$_{VP}$ washes \fbox{his}$_j$ car]$^i$}
& {\it Peter \fbox{does}$_i$ too.}
\\
&&\\
\hline
&&\\
Configuration &
VP$^j ,\dots ,$ [$_{VP} ,\dots ,$ \fbox{VPE}$_j ,\dots ,$ ]$^i$
& $,\dots ,$ \fbox{VPE}$_i ,\dots ,$ \\
&& \\
Example & {\it I'll [help you]$^j$, if you [$_{VP}$ want me \fbox{to}$_j$]$^i$}
& {\it I'll kiss you even if you \fbox{don't}$_i$.}
\\
&&\\
\hline
&&\\
Configuration &
NP$^j ,\dots ,$ [$_{NP} ,\dots ,$ \fbox{PRO}$_j ,\dots ,$ ]$^i$
& $,\dots ,$ \fbox{PRO}$_i ,\dots ,$ \\
&& \\
Example &{\it Jon$_j$ spent [$_{NP}$ \fbox{his}$_j$ paycheck]$^i$}
& {\it but Peter saved \fbox{it}$_i$.}
\\
&&\\
\hline
&&\\
Configuration &
VP$^j ,\dots ,$ [$_{NP} ,\dots ,$ \fbox{VPE}$_j ,\dots ,$ ]$^i$
& $,\dots ,$ \fbox{PRO}$_i ,\dots ,$ \\
&& \\
Example & {\it When Harry [drinks]$_j$, I conceal }
& {\it When he gambles,} 
\\
&{\it [$_{NP}$ my belief that
he \fbox{shouldn't}$_j$]$^i$}&
{\it I can't conceal \fbox{it}$_i$.}
\\
&&\\
\hline
\end{tabular}
\end{center}
\end{small}
\caption{Ellipsis, Pronouns and sloppiness}
\end{table}
In (\ref{e31a}), the target proform is a VPE whose antecedent
contains a VPE. In other words, the source proform is a VPE. Similarly, in
(\ref{e31b}), the target proform is a pronoun with a VPE as a source
proform. Both examples have a sloppy interpretation: {\it I'll kiss
  you even if you don't want me to kiss you} (instead of {\it help
  you} in the source clause) and {\it I can't conceal the belief that
  he shouldn't gamble} (instead of {\it he shouldn't drink} in the
source clause). 

What these (and the previous) examples show, is that sloppy
interpretation is independent of whether the source and the target
proforms are VPEs or pronominal anaphors. A correct theory of sloppy
interpretation must therefore be general enough to encompass all
possible cases in a uniform way. This is predicted by our account
where sloppy interpretation follows, not from the treatment of VPE and
paycheck pronouns, but from their interaction with the parallelism
constraint. In what follows we show that our approach accounts for
examples (\ref{e31a}) and (\ref{e31b}) thus covering the four possible
configurations for sloppy interpretation illustrated in
table (1) where boxes surround the source and target proforms.
Consider  (\ref{e31a}) first. Assuming that the parallel elements
are {\it help} and {\it kiss} respectively, the parallelism constraint
requires that
\begin{center}
$
\ba{l}
h(i,you) \leftarrow wt(you,h(i,you)) = A(h) \\
k(i,you) \leftarrow P(you)) = A(k) 
\ea
$
\end{center}
Resolution of the first equation yields $\lambda R. (R(i,you)
\leftarrow wt(you,R(i,you)))$ as a possible value for $A$ so that the
value for $A(k)$ is:

\[
\lambda R. (R(i,you)\leftarrow wt(you,R(i,you))(k) = A(k)
\]

\noindent
or equivalently $k(i,you)\leftarrow wt(you,k(i,you))$. Indirectly then,
the value of $P$ is now $\lambda x. wt(i,k(i,x))$ so that the VPE
occurring in the second clause is correctly interpreted as meaning
{\it want me to kiss you}.

Similarly, the derivation for (\ref{e31b}) proceeds as follows. This
time we assume that {\it drinks} and {\it gambles} are the parallel
elements so that the parallelism constraint gives rise to the
following equations:
\begin{center}
$
\ba{l}
dk(h) \rightarrow \exists x[ bel(x, sn(dk(h))) \wedge
hide(i,x)] = A(dk)\\
gb(h) \rightarrow \exists x[P(x) \wedge
hide(i,x)] = A(gb)
\ea
$
\end{center}
Resolving the first equation yields $\lambda R. (R(h) \rightarrow
\exists x[ bel(x, sn(R(h))) \wedge hide(i,x)])$ as a possible value for
$A$. By $\beta$-reduction, the value of $A(gb)$ is then $gb(h)
\rightarrow \exists x[ bel(x, sn(gb(h))) \wedge hide(i,x)]$ and $P$ is
resolved to $\lambda y. bel(y, sn(gb(h)))$. That is the second clause
of the target utterance is interpreted as meaning {\it I can't conceal
  my belief that Harry shouldn't gamble} and thereby the pronoun {\it
  it} is sloppily resolved to {\it my belief that Harry shouldn't
  gamble}.

\section{Other sloppy constructions}
\label{s6}

\cite{Kehler:icfitcodi95} lists a number of constructions where
sloppy/strict ambiguity is possible. We now consider these cases. The
first construction involves implicit referential arguments and is
exemplified in (\ref{e61a}).
\begin{example}
\label{e61a}
Jon went to a local bar to watch the Super-bowl, and Bob did too.
{\it (Bob went to a bar local to Bob/Jon)}
\end{example} 
If these implicit arguments are taken to be present in the semantic
representation, those cases are unproblematic. For instance, the
equations for (\ref{e61a}) are:
\begin{center}
$
\ba{l}
\exists x[\txt{bar}(x) \wedge \txt{loc}(x,j) \wedge
\txt{went}(j,x,\txt{watch}(j,sb))] = A(j)
\\
R(b) = A(b)
\ea
$
\end{center}
Solving the first equation yields two solutions for $A$, namely
\begin{center}
$
\ba{l}
\{A \leftarrow
\lambda z. \exists x[\txt{bar}(x) \wedge \txt{loc}(x,z) \wedge
\txt{went}(z,x,\txt{watch}(z,sb))] \}\\
\{A \leftarrow \lambda z. \exists
x[\txt{bar}(x) \wedge \txt{loc}(x,j) \wedge
\txt{went}(z,x,\txt{watch}(z,sb))] \}
\ea
$
\end{center}
The first solution yields the
sloppy reading and the second, the strict.

Another construction listed in \cite{Kehler:icfitcodi95} is one
involving either pronominal (\ref{e62a}) or definite (\ref{e62b})
event anaphora.
\begin{example}
\label{e62a}
Jon got shot by his father. That happened to Bob too.
({\it Bob got shot by Bob/Jon's father})
\end{example} \begin{example}
\label{e62b}
Jon kissed his wife, and Bill followed his example (Dahl, 1972)
({\it Bill kissed Jon/Bill's wife})
\end{example}
Here the question arises as to how expressions such as {\it that
  happened} and {\it followed x's example} should be represented.
Without going into any details about these constructions, it seems
reasonable to assume that these expressions are essentially anaphors
in that their meaning is determined by the context. Under this
assumption, the second clause of (\ref{e62a}) and
(\ref{e62b}) is then represented as $R(b)$ and the equations for
e.g. (\ref{e62a}) are:
\begin{center}
$
\ba{l}
\txt{shot}(f(j),j) = A(j)
\\
R(b) = A(b)
\ea
$
\end{center}
As usual, resolution of the first equation gives us two solutions, one
sloppy ($\{R\leftarrow \lambda z. \txt{shot}(f(z),z)\}$) and one
strict ($\{R \leftarrow \lambda z. \txt{shot}(f(j),z)\}$).

One-anaphora also permits sloppy/strict ambiguity. 
\begin{example}
\label{e64}
Although Jon bought a picture of his son, Bill snapped one himself.
\end{example}
We can capture this by treating one-anaphora similarly to paycheck
pronouns that is, as definites whose restriction is pragmatically
given. Under this assumption, the equations for (\ref{e64}) are:
\begin{center}
$
\ba{l}
\exists x[pic\_of(x,j's\/\txt{son}) \wedge bo(j,x)] = A(j,bo)\\
\exists x[P(x) \wedge sna(b,x)] = A(b,sna)
\ea
$
\end{center}
The sloppy interpretation is given by the solution
\[
\{A\leftarrow\lambda z \lambda y. \exists x[pic\_of(x,z's\/\txt{son})
\wedge y(z,x)]\}
\]
and the strict reading by the second solution
\[
\{A\leftarrow\lambda z \lambda y. \exists x[pic\_of(x,j's\/\txt{son})
\wedge y(z,x)]\}
\]

\section{Interaction with syntax and quantification }
\label{s7}

As should be clear from the above discussion, our analysis predicts a
systematic ambiguity between strict and sloppy interpretation.
However, there are certain constraints on this ambiguity which result
from the interaction of parallelism with other linguistic phenomena. We now
consider these constraints.

\subsubsection*{Syntax}
\label{s71}

Categorial information can affect sloppy/strict ambiguity. Thus 
although (\ref{e20a}) has two readings, one strict and one sloppy,
(\ref{e20b}) can only have the strict reading. 
\begin{example}
\label{e20a}
Jon$^1$'s mother loves him$_1$ and Bill's mother does too (sloppy/strict)
\end{example}
\begin{example}
\label{e20b}
Jon's mother loves Jon and Bill's mother does too (strict only)
\end{example}
Within the standard (that is, non-coloured) HOU framework, such
examples are problematic because (\ref{e20a}) and (\ref{e20b}) have
the same semantic representation and therefore, there is no way in
which they can be distinguished. In other words, both examples will be
predicted to be sloppy/strict ambiguous.  However within the HOCU
framework, colours can be used to create an interface between semantic
construction and other levels of linguistic information. In section
\ref{s2}, we saw that pronouns are variable coloured and hence
can give rise either to a strict or to a sloppy interpretation of the
target. For full NPs, we stipulate that they be \cols -coloured. Given
this, the equations for (\ref{e20b}) are:
\begin{center}
$
\ba{l}
l(m(j_\colp),j_\cols) = A_\colbvar(j_\colp)\\
R_\cols(m(b_\colp)) = A_\colbvar(b_\colp)
\ea
$
\end{center}
Crucially, resolution of the first equation yields only one solution,
not two, namely $A_\colbvar= \lambda z. l(m(z),j_\cols)$ -- this
yields the strict reading. By contrast, the sloppy reading is ruled
out because the corresponding substitution
$\{A_\colbvar\leftarrow\lambda z. l(m(z),z)\}$ is not a unifier for
the given equations. To see this, it suffices to apply this
substitution to the right-hand side of the equation,
$A_\colbvar(j_\colp)$. This yields $\lambda z. l(m(z),z)(j_\colp)$
which by $\beta$-reduction is equivalent to $l(m(j_\colp), j_\colp)$.
But this does not unify with the left-hand side of the equation
because of the colour-clash on the second occurrence of $j$. Hence the
sloppy reading is ruled out.

\subsubsection*{Scope constraints}
\label{s72}

Quantification also constrains sloppiness. This is illustrated by
the following examples. 
\begin{example}
\label{e62c}
Every$^1$ man believes he$_1$ is a fool. No, PETER believes he is a fool.
(sloppy only) \end{example} 
\begin{example}
\label{e62d}
Jon lost a$^1$ book and never got it$_1$ back. No, Peter lost a PEN
and never got it back (sloppy only) 
\end{example}
In both (\ref{e62c}) and (\ref{e62d}), the source pronoun has a
quantified antecedent and furthermore, this quantified NP does not
scope over the target utterance. As a result, only the sloppy
interpretation is possible. The strict interpretation is ruled out
because it would involve a free variable, the variable introduced by
the source quantified antecedent and occuring in the target representation. Under our analysis, this
constraint simply follows from general constraints on substitutions
which essentially say that a free variable may never become bound and
vice-versa.  Thus for (\ref{e62c}), the parallelism constraint
requires that the following equations hold (as in DSP, when a proper name has a quantified parallel element, we use the type-raised representation of proper names to guarantee parallelism of types):
\begin{center}
$
\ba{l}
\forall x [b(x) \rightarrow bel(x,f(x))] =
A(\lambda P. \forall x [b(x) \rightarrow P(x)])\\
bel(p,f(y)) = A(\lambda P. P(p))
\ea
$
\end{center}
To obtain the strict reading, the second equation needs to be
resolved as follows:

\[
\{A \leftarrow \lambda Q.Q(\lambda z. bel(z, f(x)), \/ y \leftarrow x \}
\]

Applying this substitution to the first equation, we then get 
\begin{center}
$
\ba{ll}
A(\lambda P. \forall x [b(x) \rightarrow P(x)]) & = 
\lambda Q.Q(\lambda z. bel(z, f(x))(\lambda P. \forall x [b(x)
\rightarrow P(x)]) \\
& =  \lambda P. \forall x [b(x)
\rightarrow P(x)](\lambda z. bel(z, f(x)))
\ea
$
\end{center}
And at this point, the restrictions on well-formed substitutions
requires that $x$ be renamed. Hence $x$ can never be bound by the
source quantifier {\it every man}. 

\subsubsection*{Binding constraints}

As is illustrated by the following example, the binding constraints
also play a role in constraining strict interpretation. 
\begin{example}
\label{e63}
Mary persuaded Jon$^1$ to shave himself$_1$. No, 
Mary persuaded PETER to shave himself. (sloppy only)
\end{example}
Here only the sloppy interpretation is possible. The most obvious
explanation for the missing strict reading ({\it Peter shaved Jon}) is
that the resulting structure would violate condition A of the binding
theory according to which a reflexive must be bound in its governing
category. Now recall that we only require {\it
  discourse} pronouns to be represented by free variables. In
contrast, we take reflexive and bound pronouns to be resolved at the
sentential level and hence represented either as constants or as
bound variables. Given this, the lack of strict reading in
(\ref{e63}) above, straightforwardly follows from the parallelism
constraint. The equations are:
\begin{center}
$\ba{l}
pe(m,sh(j,j)) = A(j) \\
pe(m,sh(p,p)) = A(p)
\ea
$
\end{center}
So that the only possible solution is $\{A\leftarrow\lambda x.
pe(m,sh(x,x))\}$. In particular, the solution which would yield a
strict reading namely, $\{A\leftarrow \lambda x. pe(m,sh(x,j))\}$ is
not a unifier since it cannot solve both equations simultaneously.

\section{Comparison with other approaches}
\label{s8}

In this section, we briefly compare it with three alternative
proposals: DSP's treatment of ellipsis (because it is closely related
to our proposal); Hardt's dynamic theory of ellipsis (because it uses
a dynamic rather than a static semantics and is in this sense,
fundamentally different from our proposal) and Fiengo and May's LF
approach (because contrary to our analysis which works on flat
semantic representations, this approach involves structured objects
namely, logical forms).

Let us first examine how our proposal relates to DSP's treatment of
ellipsis. In DSP's analysis, the semantic representation of the source
clause is constrained to be equal to $R(S_1, ,\dots , S_n)$ where $R$ is the semantic representation of the target ellipsis and $S_1,
,\dots , S_n$ are the semantic representations of the source parallel
elements (that is, the elements of the source clause which have an
overt parallel counterpart in the target clause). Reformulating DSP's
analysis in a way that makes it more comparable to our proposal, we
then have that the interpretation of an elliptical clause must obey
the following constraint:
\begin{center}
$
\ba{l}
S = R(S_1) ,\dots , (S_n)  \\
T = R(T_1) ,\dots , (T_n) 
\ea
$
\end{center}
\noindent
where $R$ is the semantic representation of the target ellipsis; $S$
and $T$ represent the source and the target clause and $S_1 ,\dots$$,
S_n, T_1 ,\dots , T_n$ represent the source and the target parallel
elements respectively.

This looks a lot like our parallelism constraint. There is one
important difference though: DSP's analysis requires that the
semantically underspecified element whose value is to be determined by
HOU, be the semantic representation of a VP-ellipsis. By contrast, we
only require that this semantically underspecified element be the
semantics shared by two parallel propositions.

This difference has several important consequences.
First, it allows the treatment of deaccenting (because
the free variable $R$ may represent overt rather than elided
material). Second, it captures the extended domain of
licensing involved in both ellipsis and deaccenting context (because
the free variable need not represent an ellipsis but may extend over
overt material {\it containing} an ellipsis). Third, it enables a
general theory of sloppy identity (since sloppiness is linked to
parallelism, not just ellipsis). Fourth, it allows us to preserve the
assumption that VPEs denote properties (in DSP's analysis, the
variable representing the VPE is a relation of varying arity). This
allows compatibility with Montague-type grammars that is, grammars
where the semantic type of a constituent is defined by a mapping from
syntactic categories into types.

We now turn to Hardt's approach. In a dynamic setting, anaphors can be
viewed as denoting functions from contexts to semantic
objects. Specifically, an ellipsis can be viewed as a function from
contexts to properties. As \cite{Gardent:dsve,Hardt:diovpe96} show, this
simple observation suffices to predict sloppy/strict ambiguity: if the
antecedent of an ellipsis contains a pronoun, the value of this
pronoun in the source context may differ from its value in the target
context. \cite{Hardt:diovpe96} further shows that in a dynamic setting where
pronouns and ellipsis are uniformly treated as function from contexts
to semantic objects, the free interplay between pronouns, ellipsis and
sloppy identity discussed in section \ref{s5} simply falls out. 

There are at least two main differences between Hardt's and our
proposal. First, Hardt's proposal does not impose a parallelism
constraint on sloppy identity. This has pros and cons. On the positive
side, this means that the approach encounters no difficulty in
accounting for examples such as (\ref{e39c}), where the non-contrastive
material in the target differs from its parallel counterpart in the
source. However, it also means that sloppy identity is
unrestricted. In particular, a sloppy interpretation for (\ref{e5})
cannot be ruled out.

The second difference involves directionality. Dynamic semantics is
inherently directional in that within the semantic representation, the
term representing an anaphor must be preceded by the term representing
its antecedent. As a result, the dynamic approach correctly predicts
that (\ref{e81}) lacks a sloppy interpretation that is, the reading
{\it this year, I voted for Harry} (example from
\cite{Hardt:diovpe96}).  Note that by contrast, the HOU approach will
predict both a strict and a sloppy reading for the elliptic clause.
\begin{example}
\label{e81}
Tom$^1$ is always causing me problems. Last year, I didn't vote for him$_1$ and
Tom got mad at me. This year, I did, and HARRY got mad at me.
\end{example}
The blessing is mixed however, for consider the following example.
\begin{example}
\label{e81a}
Tom is always causing me problems. Last year, he got mad at me because
I didn't vote for him and THIS year,
HARRY got mad at me because I did.
\end{example}

In this case, {\it Harry} does precede the ellipsis and consequently,
the dynamic approach predicts a sloppy reading. However, just as in
(\ref{e81}), this reading is simply not there.  In short, I don't
think example (\ref{e81}) is a very decisive argument in favour of the
dynamic approach. As example (\ref{e81a}) shows, the lack of sloppy
reading is due not to linear order, but to some other factor, possibly
the semantics of the contrast relation which holds between target and
source or possibly, the fact that {\it Tom} is the current discourse
topic.

Finally, we consider \cite{FieMay:iai94}'s account (henceforth, FM).
Under this account, ellipsis is taken to involve syntactic
reconstruction and furthermore, different types of NPs are represented
differently. For instance, a proper name is always represented by an
$\alpha$-occurrence but a pronoun can be represented either as a
$\beta$- or as an $\alpha$-occurrence.
Importantly, reconstruction is sensitive to the $\alpha -
\beta$ distinction so that specific constraints on reconstruction can
be stated which in essence, aim at capturing the effect of binding on
ellipsis reconstruction. In particular, FM's approach predicts the
difference between full NPs and pronouns discussed in section
\ref{s7}.  It also accounts for the fact that example (\ref{e82})
lacks a strict-sloppy reading that is, the reading {\it Peter said Max
saw Peter's mother}.
\begin{example}
\label{e82}
Max$^1$ said he$_1$ saw his$_1$ mother and Peter$^2$ did too
\end{example}
The HOCU approach presented here cannot explain this missing reading
for the simple reason that Co-indexed pronouns have identical
representations and are thus handled identically. Although I do no
think the problem unsolvable, I will leave it for now as an open
issue.
 
\section{Conclusion}
\label{s9} 

We have presented a uniform treatment of sloppy interpretation which
is general enough to cover the cases of sloppy/strict ambiguity
observed in the literature. Moreover the treatment integrates a
parallelism constraint which restricts sloppy interpretation to those
cases where the antecedent of the source proform has a parallel
counterpart in the target clause. 

There are two obvious directions for further research. The first
direction concerns the determination of parallelism. A real weakness
of the present paper is that parallel elements are taken as given.
This undermines the predictive power of the approach in that there is
a definite leeway in deciding what the parallel elements can or may
be. Roughly, there are two types of approach to this problem. Either,
the parallel elements are determined through some general constraint
on parallelism and contrast (cf. e.g. \cite{Hobbs:lac90}), or they are
defined through some matching mechanism on semantic representations
such as priority union (cf. e.g.
\cite{PruSchBer:dgavpa94,GroBreManMoe:puagidg94}). Both have their
merits but it seems fair to say that while the general discourse level
approach is too vague to be useful for our purpose, the more precise
approach advocated by the matching proposals is too restrictive to be
empirically very appropriate. To attain a theory of parallelism
structuring, an integration of both methods is called for, which
combines the general insights of discourse theories of parallelism
with the precision of a given matching operation.

The second line of research which needs pursuing concerns the
interaction of sloppy/strict ambiguity with other linguistic
phenomena. In section \ref{s7}, we have seen that colours allow for
the integration of non-semantic information into the semantic
construction process.  However, we have also seen cases in which
colours are not enough (cf. section \ref{s8}). In particular, our
analysis fails to predict the `eliminative puzzles of ellipsis'. That
is, it fails to capture the fact that in discourses with $n$ referring
elements in the source, less that 2$^n$ readings are actually
available (cf.  \cite{Dahl:oscsi72,FieMay:iai94}) and some  
readings are ruled out.  For these cases and more generally, for a
full treatment of how semantic construction interacts with other
levels of linguistic information, a more sophisticated apparatus is
needed. This is the subject of another paper.

\subsection*{Acknowledgments}

I would like to thank Daniel Hardt for comments and criticisms on
previous versions of this paper. The work reported here was funded by
the Deutsche Forschungsgemeinschaft in Sonderforschungsbereich
SFB--378, Project C2 (LISA).

\end{document}